\documentclass[11pt,a4paper]{article}
\usepackage[acceptedWithA]{tacl2021v1}
\usepackage{times}
\usepackage{latexsym}
\usepackage{url}
\usepackage[T1]{fontenc}
\usepackage{graphicx}
\usepackage{multicol}
\usepackage{multirow}
\usepackage{amsfonts}
\usepackage{enumitem}
\usepackage{color}
\usepackage{colortbl}

\usepackage{tikz}
\usepackage{amsmath}
\usepackage{booktabs}
\usepackage{array}
\usepackage{makecell}
\usepackage{tabu}
\usepackage{wrapfig}
\usepackage{tabularx}
\usepackage{arydshln}

\usepackage{xspace,mfirstuc,tabulary}

\title{The Impact of Word Splitting on the 
Semantic Content \\ 
of Contextualized Word Representations} 

\author{
  Aina Garí Soler$^1$ 
  \And
  Matthieu Labeau$^1$
  \\
  $^1$LTCI, Télécom-Paris, Institut Polytechnique de Paris, France, $^2$INRIA, Paris
  \\  \texttt{\{aina.garisoler,matthieu.labeau\}@telecom-paris.fr,}
  \\  \texttt{chloe.clavel@inria.fr}
  \And
  Chloé Clavel$^2$
}

\date{}

\begin{document}
\maketitle

\begin{abstract}
When deriving contextualized word representations from language models, a decision needs to be made on how to obtain one for out-of-vocabulary (OOV) words that 
are segmented into subwords. 
What is the best way to represent these words with a single vector, and are these  
representations of worse quality than those of in-vocabulary words? 
We 
carry out an intrinsic evaluation of embeddings from different models 
on semantic similarity tasks involving 
OOV words. 
Our 
analysis reveals, among other interesting findings, that the quality of representations of words that are split 
is often, but not always, worse than that of the embeddings of known words. 
Their similarity values, however, must be interpreted with caution.
\end{abstract}

\section{Introduction} 

With the appearance of 
pre-trained language models (PLMs) such as BERT \cite{devlin2019bert} and RoBERTa \cite{liu2019roberta}, 
there has been an interest in extracting, analyzing, and using contextualized word representations derived from these models, for example to understand how well they represent the meaning of words \cite{gari-soler-etal-2019-word} or to predict diachronic semantic change \cite{giulianelli-etal-2020-analysing}.

Most modern PLMs, however, operate at the subword level --- 
they rely on 
a subword tokenization 
algorithm to represent their input, like WordPiece \cite{schuster2012japanese,wu2016googles} or Byte Pair Encoding (BPE) \cite{sennrich-etal-2016-neural}. This way of representing words has advantages: with a fixed, reasonably-sized vocabulary, models can account for 
out-of-vocabulary 
words by 
splitting them into smaller units. 
When it comes to obtaining representations for words, 
a subword vocabulary implies that not all words are created equally. Words that have to be split (``split-words'') need a special treatment, different from words that have a dedicated embedding (``full-words'').

\begin{figure}
    \centering
    \includegraphics[width=\columnwidth]{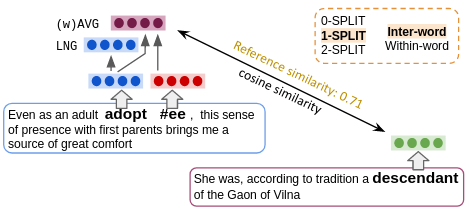} 
    \caption{Example of one of our settings 
    where we calculate the cosine similarity between the representations of an OOV word and a known word. 
    We test different ways of creating one  
    embedding for an OOV 
    word (\S \ref{sec:experimental_setup}), such as {\tt AVG} and {\tt LNG}, on two similarity tasks (\S \ref{sec:data}).} 
    \label{fig:firstpage_example}
\end{figure}

There are reasons to believe that the semantics of split-words is more poorly represented than that of full-words. First, it is generally assumed that longer tokens 
tend to contain more semantic information about a word \cite{church2020emerging} because they are more discriminative. 
The 
subword representations making up split-words must be able 
to encode the semantics of all words they can be part of. 
It has also been noted that tokenization algorithms tend to split words in a way that disregards language morphology 
\cite{hofmann-etal-2021-superbizarre}, and some of them favor splittings with more subword units than would be necessary \cite{church2020emerging}. 
In fact, 
a more morphology-aware segmentation seems to correlate with better results on downstream NLP tasks \cite{bostrom-durrett-2020-byte}.

In this study, we investigate the impact that 
word splitting (and how we decide to deal with it) has on the quality of contextualized word representations. 
We rely on the task of lexical semantic similarity estimation, which has traditionally been used as a way of intrinsically evaluating different types of word 
representations \cite{landauer1997solution,hill-etal-2015-simlex}. 
We set out to answer two main questions:

\vspace{-2mm}

\begin{itemize}
    \item What is the best strategy to combine contextualized subword representations into a contextualized word-level representation? 
\item (Given a good strategy), how does the quality of 
split-word representations compare to that of 
full-word representations? 
\end{itemize}

\vspace{-2mm}

We design experiments that allow us to answer these and related questions for BERT and other English models. 
Contrary to previous work where the quality of the lexicosemantic knowledge encoded in word representations is analyzed regardless of the words' tokenization \cite{wiedemann2019does,bommasani2020interpreting,vulic-etal-2020-probing}, we analyze the quality of the similarity estimations for split- and full-words separately, and do so in an inter-word and a within-word\footnote{Following \citet{liu-etal-2020-towards-better}'s terminology.} similarity setting. 
See Figure \ref{fig:firstpage_example} for an example of an 
experimental setting we consider. 
We uncover several interesting, and sometimes unexpected, tendencies: for example, that when it comes to polysemous nouns, OOV words 
are better represented than in-vocabulary ones;
and that 
similarity values between two split-words are generally higher than 
between two full-words.
We additionally contribute 
a new WordNet-based word similarity dataset with a 
large representation of split-words.\footnote{\url{https://github.com/ainagari/splitsim}}

\section{Background} \label{sec:background}

Subword tokenization algorithms were first proposed by \citet{schuster2012japanese} and became widespread after 
the adaptation of 
Byte Pair Encoding 
to word segmentation \cite{gage1994new,sennrich-etal-2016-neural}.
Given a specified vocabulary size, these algorithms 
create a vocabulary such that 
the most frequent character sequences in a given corpus can be represented with a single token. Unambiguous detokenization 
(i.e., recovering the original 
sequence) can be ensured 
in different ways. For example, when BERT's 
tokenizer 
splits an unknown word into multiple subwords, all 
but the first are marked with   ``\#\#'' --- we will refer to these as ``sub-tokens'' (as opposed to ``full-tokens'' which do not start with ``\#\#'').

Subword tokenization 
presented itself as a good compromise between character-level and word-level models, balancing the trade-off between vocabulary size and sequence length.
Character-based representations 
are generally better than subword-based models 
 at morphology, part-of-speech (PoS) tagging, and at handling noisy input and 
out-of-domain words; but the latter are generally better at handling semantics and syntax \cite{keren2022breaking,durrani-etal-2019-one,li-etal-2021-char}. Because of these advantages, most modern PLMs rely on subword tokenization: BERT 
uses Wordpiece; 
RoBERTa, 
XLM \cite{conneau2019cross}, GPT-2 \cite{radford2019language} and GPT-3 \cite{brown2020language} use BPE 
or some variant; T5 \cite{raffel2020exploring} relies on SentencePiece \cite{kudo-richardson-2018-sentencepiece}.   

Several works have 
pointed out that splitting words may be detrimental for certain tasks, especially if segmentation is not done in a linguistically correct way. \citet{bostrom-durrett-2020-byte}  compare two subword tokenization algorithms, BPE and 
unigramLM \cite{kudo-2018-subword}, and find that the latter, which aligns better with morphology, also yields better results on question answering, textual entailment, and named entity recognition. 
Work on machine translation has 
shown 
benefits from 
using linguistically-informed tokenization 
\cite{huck-etal-2017-target,mager-etal-2022-bpe} as well as algorithms that favor segmentation into fewer tokens 
\cite{galle-2019-investigating}. 
In fact, \citet{rust-etal-2021-good} note that multilingual BERT's (mBERT) tokenizer segments much more in some languages than others, 
and 
they demonstrate that a dedicated monolingual tokenizer plays a crucial role in mBERT's performance on numerous NLP tasks. 
Similarly, \citet{mutuvi-etal-2022-fine} 
show that increased fertility (i.e., the average number of tokens generated for every word) and number of split-words 
correlate negatively with mBERT's performance on epidemiologic watch through multilingual event extraction. 
However, the effect that (over)splitting words ---or doing so disregarding 
their morphology--- 
 has
on similarity 
remains unclear. 

 \citet{nayak-etal-2020-domain} explore a similar question to ours using the BERT model, but compare the similarity between a word representation and their 
 sub-token counterpart (e.g., \textit{night} with \textit{\#\#night}). We argue, however, that 
 even if they represent the same string, sub-tokens and full-tokens have  
 different distributions and the similarity between them is not necessarily expected to be high.\footnote{For example, in 
 \textit{hitchhiking} (tokenized $\{$\textit{hitch}, \textit{\#\#hi}, \textit{\#\#king}$\}$,  
 \textit{\#\#king} is not semantically related to 
 the word \textit{king}.} 
 Their experiments additionally involve a modification of the tokenizer.
 We instead compare representations of whole words using the models' default tokenization, 
 and we work with representations of words extracted from sentential contexts and not in isolation.
 
Multiple approaches have been proposed to improve on 
the weak aspects 
of vanilla 
subword tokenization, such as the representation of rare, out-of-domain, or misspelled words \cite{schick2020rare,hong-etal-2021-avocado, benamar-EtAl:2022:LREC}; 
and its 
concurrence with 
morphological structure \cite{hofmann-etal-2021-superbizarre}.
\citet{hofmann-etal-2022-embarrassingly} devise FLOTA, a simple segmentation method that can be used with pre-trained models without the need of re-training a new model or tokenizer. 
It consists in
segmenting words 
prioritizing the longest substrings available, omitting part of the word 
in some cases. FLOTA was shown to match the actual morphological segmentation of words more closely than the default BERT, GPT-2 and XLNet tokenizers, and yielded an improved performance on a topic-based text classification task. 
\citet{el-boukkouri-etal-2020-characterbert} propose CharacterBERT, a modified BERT model with a character-level CNN intended for building representations for complex tokens. The model improves BERT's performance on several tasks on the medical domain.
We test the FLOTA method and the CharacterBERT model in our experiments 
to investigate their advantages when it comes to 
lexical semantic similarity.

The split-words in our study are existing words ---we do not include misspelled terms--- with a generally low frequency. There has been extensive work in NLP focused on improving representations of rare words, 
which are often involved in lower quality 
predictions than those of more frequent words \cite{luong-etal-2013-better,bojanowski2017enriching,herbelot-baroni-2017-high,prokhorov2019unseen}, also in BERT \cite{schick2020rare}. Our goal is not to study the quality of rare word representations per se, but rather the effect of the splitting procedure on the quality of similarity estimates. Given the strong link between splitting and frequency, we also include an analysis controlling for this factor.

\section{Similarity Tasks and Data} \label{sec:data}

We evaluate the representations' lexical semantic content on two similarity tasks. 
In this section we describe the creation of an inter-word similarity dataset (\S 
\ref{sec:interword_dataset}) as well as the dataset used in our within-word similarity experiments (\S \ref{sec:withinword_data}). 

\subsection{Inter-Word: the {\sc split-sim} Dataset} \label{sec:interword_dataset}

We want a dataset annotated with inter-word similarities which allows 
us to compare similarity estimation quality 
in three different scenarios: when no word in a pair 
is split (\textsc{0-split}), when only one word in a pair is split (\textsc{1-split}), and when the two words 
are split (\textsc{2-split}). We refer to these 
situations, 
defined according to a given tokenizer, 
as ``split-types''.

\paragraph{Factors affecting similarity} 
 It is well known that, even in out-of-context (OOC) settings (i.e., when comparing word types and not word instances), BERT similarity predictions are more reliable when obtained from a context instead of in isolation \cite{vulic-etal-2020-probing}. However, as shown in \citet{gari-soler-apidianaki-2021-lets}, 
 representations reflect 
 the sense distribution found in the contexts used as well as the words' degree of polysemy. 
 Additionally, it is desirable to take 
 PoS into account,  
 because the quality of similarities obtained with BERT varies across PoS 
 \cite{gari-soler-etal-2022-one}. 
 To control for all these factors affecting 
 similarity estimates, we conduct separate analyses for words of different nature: monosemous nouns ({\sc m-n}), monosemous verbs ({\sc m-v}), polysemous nouns ({\sc p-n}) and polysemous verbs ({\sc p-v}). The number of senses of a word with a specific PoS is determined with WordNet \cite{Fellbaum1998}.

\paragraph{Limitations of existing datasets}
Existing 
context-dependent (i.e., not OOC) inter-word similarity datasets, like CoSimLex \cite{armendariz-EtAl:2020:LREC} and Stanford Contextual Word Similarity (SCWS) \cite{Huang:2012:IWR:2390524.2390645} do not have a large enough representation of split-words: with BERT's default tokenization 
97\% and 85\% of inter-word pairs, respectively, are of type {\sc 0-split}. 
OOC word similarity datasets 
do not meet our criteria either. 
In Simlex-999 \cite{hill-etal-2015-simlex} and WS353 \cite{agirre-etal-2009-study}, 96\% and 95\% pairs are 
\textsc{0-split}. 
CARD-660 \cite{pilehvar-etal-2018-card}, which specifically targets rare words, has a better distribution of split-types, but it
contains a large number of multi-word expressions (MWEs) and lacks PoS information. 
The Rare Word (RW) dataset \cite{luong-etal-2013-better} is also specialized on rare words and has a larger coverage of {\sc 1-} and {\sc 2-split} pairs, but 
we do not use it because of its low inter-annotator agreement and problems with 
annotation consistency described in \citet{pilehvar-etal-2018-card}.

Therefore, and since it is 
more convenient 
to obtain similarity annotations out-of- rather than in-context, 
we create a dataset of OOC word similarity, {\sc split-sim}. It consists of four separate subsets, one for each type of word. 
Each subset 
has a balanced representation of split-types. 

\paragraph{Word selection and sentence extraction} We use WordNet 
to create 
{\sc split-sim}. 
We first identify all 
words  
in WordNet which are not MWEs, numbers or proper nouns, 
and which are at least two characters long. After this filtering, we find 28,563 monosemous nouns, 12,903 polysemous nouns, 3,888 monosemous verbs and 4,518 polysemous verbs.

We search for 
sentences containing these words 
in the c4 corpus \cite{raffel2020exploring}, from which we will derive 
contextualized word representations. 
We postag sentences using \texttt{nltk} \cite{bird2009natural}.\footnote{{\tt nltk} offers a good speed/accuracy trade-off 
compared to SpaCy, Flair \citep{akbik-etal-2019-flair}, stanza 
\citep{qi2020stanza} and the RDRPOSTagger \citep{nguyen-etal-2014-rdrpostagger}. The agreement between the {\tt nltk} and SpaCy tags for the target words in our final set of selected sentences is of 89.8\%.} 
Importantly, we only select sentences that contain the lemma form of a word with the correct PoS. This ensures that a word will be tokenized in the same way (and belong to the same split-type) in all its contexts, and 
avoids 
BERT's word form bias \cite{laicher-etal-2021-explaining}. 
We only keep 
words for which we could find at least ten sentences that are between 5 and 50 words long.
If we found more, we randomly select 10 sentences among the first 100 occurrences found.

\paragraph{Pair creation}  
We rely on \textsc{wup} 
\cite{wu-palmer-1994-verb}, a Wordnet-based similarity measure, as our reference similarity value. 
\textsc{wup} similarity takes into account the depth (the path length to the root node) of the two senses to be compared (s$_1$ and s$_2$), as well as of their ``least common subsumer'' (LCS). In general, the deeper LCS is, the higher the similarity between s$_1$ and s$_2$.\footnote{Since {\sc wup} is a \underline{sense} similarity measure, 
we define the similarity of two polysemous words to be the highest similarity found between all possible pairings of their senses.}

\begin{table}[]
    \centering
    \scalebox{0.95}{
    \begin{tabular}{lc|cc}
    \textbf{Dataset} & \textbf{PoS} & $\rho$   & \textbf{\# pairs} \\
    \hline
    \multirow{2}{*}{Simlex-999} & n & 0.55 & 666  \\
     & v & 0.39 & 162 \\
     \hline
     \multirow{2}{*}{WS353} & n & 0.64 & 201 \\
     & v & 0.10 & 29 \\
     \hline
     \multirow{2}{*}{CARD-660} & n & 0.64 & 170 \\
     & v & 0.50 & 20 \\
     \hline
     \multirow{2}{*}{RW} & n & 0.24 & 910 \\
     & v & 0.25 & 681 \\    
    \end{tabular}}
    \caption{Spearman's $\rho$ between {\sc wup} similarity and human judgments from existing word similarity datasets.}
    \label{tab:correlation_wup_manual}
\end{table}

{\sc wup} similarities are only available for nouns and verbs. It is important to note that 
similarities for the two PoS follow slightly different distributions, which is another reason for keeping them separate.
We choose {\sc wup} 
over other WordNet-based similarity measures like {\sc lch} \cite{leacock-etal-1998-using} 
and path similarity because it conveniently ranges from 0 to 1 and 
its distribution aligns with the intuition that most randomly obtained pairs would have a low semantic similarity.\footnote{We observed the distribution of similarity values of the three measures on a random sample of 2,000 
lemmas. Similarities are calculated using {\tt nltk}.} 
{\sc wup} is not as good as human judgments, but it correlates reasonably well with them \cite{yang2019semantic}. Table \ref{tab:correlation_wup_manual} shows the measure's correlation with manual similarity judgments by PoS. 
We 
consider it to be a good enough approximation for our purposes of 
comparing performance across split-types and representation strategies. 
For an alternative non-Wordnet-based similarity metric to compare to {\sc wup}, we also use the similarity of FastText embeddings \cite{bojanowski2017enriching} as a control. 

\begin{table}[]
    \centering
    \scalebox{0.87}{
    \begin{tabular}{lll|cccc}
         &&& {\sc m-n} & {\sc m-v} & {\sc p-n} & {\sc p-v}  \\
         \hline
         \multirow{7}{*}{\rotatebox[origin=c]{90}{full}} & \multirow{3}{*}{\rotatebox[origin=c]{90}{BERT}} & {\sc 0-split} & 22,500 & 850 & 5,000 & 5,000 \\
         &&{\sc 1-split} & 22,500 & 850 & 5,000 & 5,000 \\
         &&{\sc 2-split} & 22,500 & 850 & 5,000 & 5,000 \\         
         \cdashline{2-7}
         & \multirow{3}{*}{\rotatebox[origin=c]{90}{XLNet}} & {\sc 0-split} & 12,166 & 644 & 3,642 & 5,610 \\
         && {\sc 1-split} & 25,490 & 1,033 & 6,009  & 6,006 \\
         && {\sc 2-split} & 29,844 & 873 & 5,349 & 3,384 \\        
         \cdashline{2-7}
         && \textbf{Total} & 67,500 & 2,550 & 15,000 & 15,000 \\
         \hline
         \multirow{7}{*}{\rotatebox[origin=c]{90}{balanced}} & \multirow{3}{*}{\rotatebox[origin=c]{90}{BERT}} & {\sc 0-split} & 7,387 & 122 & 572 & 240 \\
         &&{\sc 1-split} & 3,873 & 119 & 973 & 687 \\
         &&{\sc 2-split} & 1,915 & 146 & 1,553 & 1,776 \\                  
         \cdashline{2-7}
         & \multirow{3}{*}{\rotatebox[origin=c]{90}{XLNet}} & {\sc 0-split} & 2,491 & 74 & 317 & 563 \\
         && {\sc 1-split} & 5,992 & 165 & 1,149  & 1,270 \\
         && {\sc 2-split} &  4,692 & 148 & 1,632 & 870 \\
         \cdashline{2-7}
         && \textbf{Total} & 13,175 & 387 & 3,098 & 2,703 \\
         \hline         
    \end{tabular}}
    \caption{Composition of the {\sc split-sim} dataset (full and balanced versions) according to two different tokenizers.}
    \label{tab:splitsim_composition}
\end{table}

We exhaustively pair all words in each subset and calculate their {\sc wup} similarity. We select a portion of all pairs ensuring that the full spectrum 
of similarity values is represented: 
For each split-type, we randomly sample the same number of word pairs in each 0.2-sized similarity score interval. Due to data availability this number is different for each subset. 
For the creation of the dataset, the split-type is determined using BERT's default tokenization. Table \ref{tab:splitsim_composition} contains statistics on the full dataset composition. 
Example pairs from the dataset can be found \mbox{in Table \ref{tab:examples_dataset}.}

\begin{table}[t!]
    \centering
    \scalebox{0.93}{
    \begin{tabular}{ll!{\color{black}\vrule}c!{\color{black}\vrule}c}
     \multicolumn{2}{c|}{\textbf{Word pairs}} & \textbf{Split-type} & {\sc wup} \\
     \hline     
         \{accordion\} & \{guitar\} & {\sc 0-split} & 0.80  \\ 
         \arrayrulecolor{lightgray}\hline
         \begin{tabular}{@{}c@{}} \{tom, \#\#fo, \\ \#\#ole, \#\#ry\}\end{tabular} &  \{loaf, \#\#ing\}    & {\sc 2-split}  & 0.63 \\ 
         \arrayrulecolor{lightgray}\hline
         \{ethanol\}  & \{fuel\}   & {\sc 0-split} & 0.46 \\ 
         \arrayrulecolor{lightgray}\hline
         \{ash, \#\#tray\} & \{weather\}  & {\sc 1-split} & 0.24 \\         
    \end{tabular}}
    \caption{Example word pairs from {\sc split-sim} ({\sc m-n} subset) with their BERT tokenization.}
    \label{tab:examples_dataset}
\end{table}

\begin{table}[t]
    \centering
    \scalebox{1}{
    \begin{tabular}{ll|cccc}
         & & {\sc m-n} & {\sc m-v} & {\sc p-n} & {\sc p-v}  \\
         \hline
         \multirow{3}{*}{\rotatebox[origin=c]{90}{full}} & {\sc 0-split} & 3.75 & 3.99 & 4.09 & 4.30 \\
        & {\sc 1-split} & 2.66 & 2.93 & 3.15 & 3.27 \\
         &{\sc 2-split} & 1.54 & 1.81 & 2.18 & 2.25 \\
         \hline
         \multirow{3}{*}{\rotatebox[origin=c]{90}{\small{balanced}}} &{\sc 0-split} & 3.35 & 3.38 & 3.43 & 3.51 \\
         &{\sc 1-split} & 3.04 & 3.09 & 3.14 & 3.19 \\
         &{\sc 2-split} & 2.72 & 2.81 & 2.84 & 2.90 \\

    \end{tabular}}
    \caption{Average frequencies 
    in each {\sc split-sim} subset 
    (BERT tokenization). Values are the base-10 logarithm of the number of times a word appears per billion words. For reference, the frequencies of \textit{can}, \textit{dog}, \textit{oatmeal} and \textit{myxomatosis} are 6.46, 5.10, 3.37 and 1.61.}
    \label{tab:freq_in_splitsim}
\end{table}

\paragraph{Controlling for frequency}
In our experiments we also want to control for frequency, since split-words tend to be more rare than full-words. We calculate the frequencies of 
words in {\sc split-sim} with 
the \texttt{wordfreq} Python package \cite{robyn_speer_2022_7199437} and report them in Table \ref{tab:freq_in_splitsim}.   
Frequencies are low overall, especially those of monosemous split-words. 
To mitigate the potential effect of frequency differences, we find the narrowest possible frequency range that is still represented with enough word pairs in every split-type. We determine this range to be [2.25, 3.75). We create a smaller version of {\sc split-sim}, which we call ``balanced'', with pairs that include only words within this frequency interval.
Another aspect to take into account is that of the difference in frequencies of words in a pair, what we call $\Delta f$. 
$\Delta f$ is highest in {\sc 1-split} pairs (up to 2.19 in {\sc m-v} compared to 0.67 in the corresponding {\sc 0-split}), but it is much lower overall 
in the balanced dataset because of the narrower frequency range.

\begin{table}[]
    \centering
    \scalebox{0.79}{
    \begin{tabular}{ll|c|ccc}
    & & \textbf{Training} & \multicolumn{3}{c}{\textbf{Evaluation}} \\
    \hline
    & & {\sc 0-split} & {\sc 0-split} & {\sc 1-split} & \textsc{2-split} \\
    \hline
    \hline

\multirow{6}{*}{\rotatebox[origin=c]{90}{BERT}} & \textbf{All} & 5,104  & 479  & 117  & 366  \\
& {\sc same} & 3,388  & 312  & 0 & 274  \\
& {\sc diff} & 1,716  & 167  &  117 & 92 \\
& \textbf{T} & 2,464 & 228 & 72 & 269 \\
& \textbf{F} &  2,640 & 251 & 45 & 97 \\
\cdashline{2-6}
& \textbf{Lemmas} & 1,043 & 445  & 102  & 288  \\
\hline
\multirow{6}{*}{\rotatebox[origin=c]{90}{XLNet}} & \textbf{All} & 4,648  & 415  & 502 & 501  \\
& {\sc same} & 3,144  & 292  & 142 & 396  \\ 
& {\sc diff} & 1,504  & 123  & 360 & 105 \\ 
& \textbf{T} & 2,272 & 203 & 222 & 336 \\ 
& \textbf{F} &  2,376 & 212 & 280 & 165 \\ 
\cdashline{2-6}
& \textbf{Lemmas} & 944 & 387  & 291  & 351  \\
\hline
    \end{tabular}}
    \caption{WiC statistics: number of word pairs of different types and number of unique lemmas with different tokenizers.}
    \label{tab:wic_stats}
\end{table}

\subsection{Within-Word} \label{sec:withinword_data}

Similarly to the inter-word setting, for within-word similarity we want to distinguish between {\sc 0-}, {\sc 1-} and {\sc 2-split} pairs. 
An 
important factor that can influence within-word similarity estimations is whether pairs compare the 
same word form ({\sc same}) or different morphological forms of the word ({\sc diff}). {\sc 1-split} pairs are all necessarily of type {\sc diff},\footnote{Except for XLNet, which is a cased model.} 
but 
{\sc 0-} and {\sc 2-split} pairs can be of either type 
(e.g., \{\textit{carry}\} vs \{\textit{carries}\}; \{\textit{multi}, \textit{\#\#ply}\} vs
\{\textit{multi}, \textit{\#\#ply}, \textit{\#\#ing}\}). 

We choose the Word-in-Context (WiC) dataset \cite{pilehvar-camacho-collados-2019-wic}
for its convenient representation of all split-types. 
WiC contains pairs of word instances that have the same (T) or a different (F) meaning. 
We use the training and development sets, 
whose labels (which are taken as a reference) are publicly available. 
They consist of a total of 6,066 pairs that we rearrange for our purposes. 
We use as training data all {\sc 0-split} pairs 
found in the original training set. For evaluation we use the 
{\sc 0-split} pairs in the original development set, and all  {\sc 1-split} and {\sc 2-split} pairs 
found in both sets. 
Table \ref{tab:wic_stats} contains 
details about the composition of the dataset, such as the 
proportion of T and F labels. Note that, again, numbers differ depending on the tokenizer used (BERT's or XLNet's).

WiC is smaller than {\sc split-sim} and offers a less controlled, but more realistic, environment. For example, 
{\sc 2-split} pairs 
involve words with low frequency and few senses, which 
results in an overrepresentation of T pairs in this class. 
We did not use other 
within-word similarity datasets such as Usim \cite{erketal2009,erk2013measuring} or 
DWUG \cite{schlechtweg-etal-2021-dwug}, because they contain a small number of {\sc 1-} and {\sc 2-split} pairs (91 and 4 in Usim), or these 
involve very few distinct 
lemmas (14 and 12 in DWUG).

\section{Experimental Setup} \label{sec:experimental_setup}

\subsection{Models}

We run all our experiments with representations extracted from the BERT (base, uncased) 
model in the \texttt{transformers} library \cite{wolf-etal-2020-transformers} and the \texttt{general} CharacterBERT model (hereafter CBERT).\footnote{\url{https://github.com/helboukkouri/character-bert}} The two are trained on a comparable amount of tokens (3.3B and 3.4B, respectively) which include English Wikipedia. BERT is also trained on BookCorpus \cite{zhu2015aligning}, and 
CBERT 
on OpenWebText \cite{openwebtext}. 
For comparison, we also include ELECTRA base \cite{clark2020electra} and XLNet (base, cased)\footnote{The cased and uncased versions of a word may be split differently. To avoid inconsistencies in the definition of split-types in {\sc split-sim}, 
target words are presented in lower case exclusively.} 
 \cite{yang-2019-xlnet} in our analysis. ELECTRA is trained on the same data as BERT and uses exactly the same architecture, tokenizer and vocabulary (30,522 tokens), but is trained with a more efficient discriminative pre-training approach. XLNet relies on the SentencePiece 
implementation of UnigramLM %
and has a 32,000 token vocabulary. It is a Transformer-based model pre-trained on 32.89B tokens with the task of Permutation Language Modeling. We choose these models because they are newer and better than BERT (e.g., on GLUE \cite{wang-etal-2018-glue} among other benchmarks) and because of their wide use. 
XLNet allows us to investigate the effect of word splitting in models relying on different tokenizers. We experiment with all layers of the models.
In inter-word experiments, a word representation is obtained by averaging the contextualized word representations from each of the 10 sentences.

\begin{table*}[!t]
    \centering
    \scalebox{0.86}{
    \begin{tabular}{l| ccc| ccc| c | ccc | ccc}    
    \multicolumn{1}{c|}{} & \multicolumn{3}{c|}{BERT} & \multicolumn{3}{c|}{BERT-FLOTA} & \multicolumn{1}{c|}{CBERT} & \multicolumn{3}{c|}{ELECTRA} & \multicolumn{3}{c}{XLNet}  \\
    
    & {\tt AVG} & {\tt WAVG} & {\tt LNG} & {\tt AVG} & {\tt WAVG} & {\tt LNG} & - & {\tt AVG} & {\tt WAVG} & {\tt LNG} & {\tt AVG} & {\tt WAVG} & {\tt LNG} \\
    \hline
    {\sc m-n} & 38$_{6}$ & 38$_{6}$ & 31$_{8}$ & 35$_{5}$ & 35$_{6}$ & 30$_{8}$ & 40$_{10}$ & 39$_{5}$ & 40$_{5}$ & 35$_{5}$ & 41$_{10}$ & \textbf{42$_{4}$} & \textbf{42$_{4}$} \\
    
    {\sc m-v} & 33$_{11}$ & 33$_{11}$ & 31$_{12}$ & 27$_{12}$ & 28$_{12}$ & 25$_{12}$ & 31$_{3}$ & 34$_{5}$ & 35$_{3}$ & 28$_{5}$ & 36$_{4}$ & \textbf{37$_{4}$} & \textbf{37$_{4}$} \\
    {\sc p-n} & 33$_{10}$ & 34$_{10}$ & 29$_{12}$ & 28$_{10}$ & 28$_{10}$ & 26$_{12}$ & 29$_{10}$ & 34$_{8}$ & 35$_{6}$ & 32$_{7}$ 
    & 35$_{10}$ & 36$_{10}$  & \textbf{37$_{5}$} \\
    
    {\sc p-v} & 30$_{10}$ & 30$_{12}$ & 28$_{12}$ & 24$_{12}$ & 24$_{12}$ & 21$_{12}$ & 25$_{10}$ & 27$_{8}$ & 28$_{8}$ & 26$_{7}$ &  29$_{7}$ & 31$_{6}$ & \textbf{33$_{4}$} \\                
    \end{tabular}}
    \caption{Spearman's $\rho$ ($\times$ 100) 
    obtained on {\sc split-sim} with different representation types and strategies. Subscripts denote the best layer. The best result on each subset is boldfaced.} 
    \label{tab:global_results_splitsim}
\end{table*}

\subsection{Input Treatment} \label{sec:reptypes}

Here we describe the different ways in which 
input data is processed 
before feeding it to the models. 

\paragraph{Tokenization} We use the model's default tokenizations. We additionally experiment with the FLOTA tokenizer \cite{hofmann-etal-2022-embarrassingly} used in combination with BERT. FLOTA has a hyperparameter controlling the number of iterations, $k \in \mathbb{N}$. With lower $k$, portions of words are more likely to be omitted. We set $k$ to 3 as it obtained the best results on text classification \cite{hofmann-etal-2022-embarrassingly}.

\paragraph{Lemmatization}
In the WiC dataset, the word instances to be compared may have different surface forms. 
One way of restricting 
the influence of word form on BERT representations 
is through lemmatization \cite{laicher-etal-2021-explaining}. 
We replace the target word instance with its lemma 
before extracting its representation.  We refer to this setting as {\tt LM}. This procedure is not relevant for {\sc split-sim}, where 
all instances are already in lemma form.

\subsection{Split-Words Representation Strategy} \label{sec:splitword_strategies}

We compare different strategies for pooling 
a single word embedding 
from the representations of a split-word's multiple subwords. 

\paragraph{Average ({\tt AVG})} The embeddings of every subword forming a word are averaged to obtain a word representation. This is the most commonly used strategy when representing split-words 
\cite[inter alia]{wiedemann2019does,gari-soler-etal-2019-word,liu-etal-2020-towards-better,montariol-allauzen-2021-measure}. \citet{bommasani2020interpreting} tested {\tt max}, {\tt min} and {\tt mean} pooling as well as using the representation of the last token. We only use {\tt mean} pooling ({\tt AVG}) from their work because they found it to work best for OOC word similarity.

\paragraph{Weighted average ({\tt WAVG})} A word is represented with a weighted average of all its subword representations. Weights are assigned according to word length. 
For example, a subword that makes up 70\% of a word's characters is weighted with 0.7.

\paragraph{Longest ({\tt LNG})} Only the representation of the longest subword is used.  This approach, as {\tt WAVG}, accounts for the intuition that longer pieces carry 
more information about the meaning of a word.

\subsection{Prediction and Evaluation}
The similarity between two words or word instances is calculated as the cosine similarity between their representations. 
For experiments on {\sc split-sim}, the evaluation metric is Spearman's $\rho$. 
For within-word experiments, we train a logistic regression classifier that uses the cosine between two word instance representations as its only feature. We evaluate the classifier based on its accuracy.

\section{Results and Analysis} \label{sec:results_analysis}

In this section we analyze the results obtained on the {\sc split-sim} (\S  \ref{sec:interword_results}) and WiC (\S  \ref{sec:withinword_results}) datasets.

\subsection{Inter-Word} \label{sec:interword_results}

We start with  a look at the results of each method on each {\sc split-sim} subset as a whole. The rest of this section is 
organized 
around the main questions we aim to answer.

Table \ref{tab:global_results_splitsim} presents the correlations obtained by different representation types and strategies 
on the full dataset. We report the highest correlation found across all layers. The best model on all subsets is clearly XLNet with the {\tt LNG} or {\tt WAVG} strategies. ELECTRA (with {\tt WAVG}) is the second best one on most subsets.
Correlations obtained against FastText cosine similarities reflect, with few exceptions, the same tendencies observed in this section (results are presented in Appendix \ref{app:fasttext}). 

\begin{table*}[!t]
    \centering
    \scalebox{0.87}{
    \begin{tabular}{ll|ccc | ccc | c | ccc | ccc}    
    & \multicolumn{1}{c|}{} & \multicolumn{3}{c|}{BERT} & \multicolumn{3}{c|}{BERT-FLOTA} & \multicolumn{1}{c|}{CBERT} & \multicolumn{3}{c|}{ELECTRA} & \multicolumn{3}{c}{XLNet}  \\
    & & {\tt AVG} & {\tt WAVG} & {\tt LNG} & {\tt AVG} & {\tt WAVG} & {\tt LNG} & - & {\tt AVG} & {\tt WAVG} & {\tt LNG} & {\tt AVG} & {\tt WAVG} & {\tt LNG} \\
    \hline
    \multirow{3}{*}{{\sc m-n}} & {\sc 0-s} & \multicolumn{3}{c|}{49} & \multicolumn{3}{c|}{49} & 48 & \multicolumn{3}{c|}{48} & \multicolumn{3}{c}{\underline{51}} \\ 
    \cdashline{3-15}
    & {\sc 1-s} & \textbf{41}* & 38* & 28* & \textbf{35}* & 33* & 26* & 41* & \textbf{\underline{42}}* &  40* & 35* &     \textbf{\underline{46}}*  & \textbf{\underline{46}}* & 44* \\
    & {\sc 2-s} & \textbf{43}* & 40* & 26* & \textbf{34}* & 32* & 23* & \underline{47} & \textbf{45}* & 43* & 31* & 
    \textbf{46}* & 45* & 39* \\       
    \hline
\multirow{3}{*}{{\sc m-v}} & {\sc 0-s} & \multicolumn{3}{c|}{43} & \multicolumn{3}{c|}{43} & 39 & \multicolumn{3}{c|}{42} & \multicolumn{3}{c}{\underline{50}} \\ 
\cdashline{3-15}
    & {\sc 1-s} & \textbf{33}* & \textbf{33}* & 26* & \textbf{23}* & \textbf{23}* & 19* & 28* & \textbf{\underline{36}} & \textbf{\underline{36}} & 26* & \textbf{34}* & \textbf{34}* & 32* \\ 
    & {\sc 2-s} & \textbf{\underline{41}} & 40 & 32* & \textbf{25}* & \textbf{25}* & 23* & 35 & 36 & \textbf{38} & 28* &  \textbf{39}* & 37* & 35* \\ 
    \hline
    \multirow{3}{*}{{\sc p-n}} & {\sc 0-s} & \multicolumn{3}{c|}{\underline{38}} & \multicolumn{3}{c|}{\underline{38}} & 31 & \multicolumn{3}{c|}{32}  & 
    \multicolumn{3}{c}{\underline{38}} \\ 
    \cdashline{3-15}
    & {\sc 1-s} & \textbf{\underline{38}} & 35 & 28* & \textbf{32}* & 30* & 24* & 31 & \textbf{\underline{38}}* & 37* & 34 & 
    \textbf{39} & 38 & 37\\    
    & {\sc 2-s} & \textbf{41}* & 37 & 25* & \textbf{29}* & 27* & 20* & 43* & \textbf{\underline{45}}* & 44* & 36* & 
    \textbf{45}* & 43* & 39 \\
    \hline
\multirow{3}{*}{{\sc p-v}} & {\sc 0-s} & \multicolumn{3}{c|}{\underline{37}} & \multicolumn{3}{c|}{\underline{37}} & 31 & \multicolumn{3}{c|}{34} & \multicolumn{3}{c}{\underline{37}} \\ 
\cdashline{3-15}
    & {\sc 1-s} & \textbf{34}* & 33* & 25* & \textbf{26}* & 23* & 18* & 25* & \textbf{30}* & \textbf{30}* & 27* & 
    \textbf{\underline{35}} & 33* & 32* \\ 
    & {\sc 2-s} & \textbf{33}* & 31* & 24* & \textbf{16}* & 15* & 14* & 31 & 31* & \textbf{32} & 26* & \textbf{\underline{34}} & 33* & 31* \\  
    \end{tabular}}
    \caption{Spearman's $\rho$ ($\times$ 100) 
    on 
    {\sc split-sim} (full). 
    The best result by subset, split-type and model is boldfaced. The best overall result in every subset and split-type is underlined. * indicates that a 1- or {\sc 2-split} correlation coefficient is significantly different ($\alpha < 0.05$) from the corresponding {\sc 0-split} result  
\cite{sheskin2003handbook}.}
    \label{tab:bysplit_results_splitsim}
\end{table*}

\subsubsection{What is the best strategy to represent split-words?} \label{sec:best_strategy}

Table \ref{tab:bysplit_results_splitsim} 
shows the Spearman's correlations obtained by different 
pooling methods on the three split-types. 
The best layer is selected 
separately for each split-type, model and strategy. 
We can see that the best strategy for each model tends to be stable across datasets. {\tt AVG} is the preferred strategy overall, followed by {\tt WAVG}, which, in ELECTRA and XLNet, performs almost on par with {\tt AVG}. Using the longest subword ({\tt LNG}) results in a considerably lower performance across models and data subsets, presumably because some important information is excluded from the representation.
CBERT obtains good results (comparable or better than BERT) on monosemous nouns ({\sc m-n}), but on other kinds of words it generally lags behind. 

\begin{table}[!t]
    \centering
    \scalebox{0.8}{
    \begin{tabular}{ll|cc|cc|cc}    
    & \multicolumn{1}{c|}{} & \multicolumn{2}{c|}{{\tt AVG}} & \multicolumn{2}{c|}{{\tt WAVG}} & \multicolumn{2}{c}{{\tt LNG}} \\
    & & {\sc com} & {\sc incm} & {\sc com} & {\sc incm} & {\sc com} & {\sc incm} \\ 
    \hline
    \multirow{2}{*}{{\sc m-n}} & {\sc 1-s} & \textbf{36} & 30 & \textbf{33} & 30 & 26 & \textbf{30} \\    
    & {\sc 2-s} & 31 & \textbf{41} & 29 & \textbf{40} & 20 & \textbf{28} \\
    \hline
    
\multirow{2}{*}{{\sc m-v}} & {\sc 1-s} & 22 & \textbf{25} & \textbf{23} & 22  & \textbf{20} & 03 \\
    & {\sc 2-s} & 23 & \textbf{32} & 24 & \textbf{31} & 22 & \textbf{27} \\
    \hline
    
    \multirow{2}{*}{{\sc p-n}} & {\sc 1-s} & \textbf{33} & 22 & \textbf{30} & 29 & \textbf{24} & 21 \\    
    & {\sc 2-s} & \textbf{30} & 24 & \textbf{28} & 24 & \textbf{21} & 17  \\
    \hline
\multirow{2}{*}{{\sc p-v}} & {\sc 1-s} & \textbf{26} & 16 & \textbf{24} & 09 & \textbf{19} & -02  \\
    & {\sc 2-s} & \textbf{17} & 09 & \textbf{17} & 07 & \textbf{15} & 07 \\
    
    \end{tabular}}
    \caption{Results obtained with FLOTA tokenization on pairs where words were fully preserved ({\sc com}) and where at least one word had a portion omitted ({\sc incm}).}
    \label{tab:flota_analysis}
\end{table}

\paragraph{FLOTA performance}
The use of the FLOTA tokenizer systematically decreases BERT's performance. 
We believe there are two main reasons behind this outcome: First, that similarly to {\tt LNG}, FLOTA sometimes\footnote{With FLOTA, 
9.8 to 20.8\% of {\sc 1-} and {\sc 2-split} pairs (depending on the dataset) have at least one incomplete word.} omits parts of words.
We investigate this by comparing its performance on pairs where both words were left complete ({\sc com}) to that on pairs where some word 
is incomplete ({\sc incm}). We present results in Table \ref{tab:flota_analysis}. We observe that, indeed, in most cases, performance is worse when parts of words are omitted. However, this is not the only factor at play, since the performance on {\sc com} is still lower than when using BERT's default tokenizer. 
The second reason, we believe, is that FLOTA tokenization differs from the tokenization used for BERT's pretraining. 
FLOTA was originally evaluated on a \textit{supervised} text classification task \cite{hofmann-etal-2022-embarrassingly}, while we do not 
fine-tune the model for similarity estimation with the new tokenization. Additionally, classification was done relying on a sequence-level token representation (e.g., [CLS] in BERT). It is possible that FLOTA tokenization provides an advantage when considering full sequences which does not translate to an improvement in the similarity between individual word token representations. 
Given its poor results compared to BERT, in what follows, we omit FLOTA from our discussion.

\subsubsection{Is performance on pairs involving split-words worse than on {\sc 0-split}?} \label{sec:isperformanceworse}

In Table \ref{tab:bysplit_results_splitsim} we can 
see that, as expected, in most subsets ({\sc m-n}, {\sc m-v} and {\sc p-v}), performance is worse in pairs involving split-words. This is however not true of polysemous nouns ({\sc p-n}), where 
similarities obtained with all models are of better or comparable quality on {\sc 1-} and {\sc 2-split} pairs. With 
CBERT, performance on {\sc 2-split} pairs is never significantly lower than on {\sc 0-split} pairs.

\paragraph{Lower correlation of polysemous words} Correlations obtained on polysemous words are overall lower than on monosemous words, particularly so in the {\sc 0-split} case. Worse performance on polysemous words can be expected for two main reasons. 
First, {\sc wup} between polysemous words is determined as the maximum similarity attested for all their sense pairings, while cosine similarity takes into account all the contexts provided as well as the accumulated lexical knowledge about the word contained in the representation. Second, the specific sense distribution found in the randomly selected contexts may also have an impact on the final results (particularly if, e.g., the relevant sense for the comparison is missing).

\paragraph{{\sc 1-split} vs {\sc 2-split}} Another interesting observation is that, in most cases, performance on {\sc 1-split} pairs is lower than on {\sc 2-split} pairs. 
We identify two main factors that explain 
this result. 
One is the fact that 
in {\sc 1-split}, the words in a pair 
are represented using different strategies (the plain representation vs {\tt \{AVG|WAVG|LNG\}}). 
In fact 
exceptions to this observation concern almost exclusively 
the {\tt LNG} pooling strategy. {\tt LNG} does not involve any arithmetic operation, which makes the representations of the split- and full-word in a {\sc 1-split} pair more comparable to each other.
Another explanation is the difference in frequency between words 
($\Delta f$), which tends to be larger in {\sc 1-split} than in 
0- and {\sc 2-split} pairs. 
We explore this possibility in our frequency analysis below.

In the remaining inter-word experiments, we focus our observations on the better (and simpler) {\tt AVG} strategy.

\subsubsection{Frequency-related analysis}

As explained in Section \ref{sec:interword_dataset}, frequency and word-splitting are strongly related. The experiments presented in this section help us understand how the 
tendencies observed so far are linked to or affected by word frequency.

\begin{table}[t!]
    \centering
    \scalebox{0.86}{
    \begin{tabular}{ll|c  | c | c | c}    
    &  & BERT  & CBERT & ELECTRA & XLNet  \\    
    \hline
    \multirow{3}{*}{{\sc m-n}} & {\sc 0-s} & \textbf{52} & \textbf{52} & \textbf{53} & 
    \textbf{57} \\    
    & {\sc 1-s} & 47* & 49* & 49* & 53* \\
    & {\sc 2-s} & 44*  & 47* & 48* &  49* \\
    \hline
\multirow{3}{*}{{\sc m-v}} & {\sc 0-s} & \textbf{53} & \textbf{54} & \textbf{60} & \textbf{71} \\ 
    & {\sc 1-s} & 39 & 32* & 31* & 46* \\
    & {\sc 2-s} & 42 & 32* & 40* & 36* \\
    \hline
\multirow{3}{*}{{\sc p-n}} & {\sc 0-s} & 39 & 41 & 44 & 47 \\ 
    & {\sc 1-s} & \textbf{45} & \textbf{46} & \textbf{46} & \textbf{48} \\
    & {\sc 2-s} & 41 & 40 & 44 & 42 \\
    \hline
\multirow{3}{*}{{\sc p-v}} & {\sc 0-s} & \textbf{46} & \textbf{46} & \textbf{46} & \textbf{48} \\ 
    & {\sc 1-s} & 39 & 37 & 44 & 46  \\
    & {\sc 2-s} & 39 & 35* & 40 &  40* \\  
    
    \end{tabular}}
    \caption{Spearman’s $\rho$ (× 100) on 
    {\sc split-sim} (balanced), {\tt AVG} strategy. The best result by subset and model
is boldfaced. 
* indicates that a 1- or {\sc 2-split}
correlation coefficient is significantly different 
from the corresponding {\sc 0-split} result \cite{sheskin2003handbook}.}   \label{tab:bysplit_results_splitsim_balanced}
\end{table}

\paragraph{Controlling for frequency} 
The lower correlations obtained in {\sc 1-} and {\sc 2-split} pairs in most subsets could simply be due to the lower frequency of split-words, and not necessarily to the fact that they are split.
To verify this, we evaluate the models' predictions on word pairs found in the balanced 
{\sc split-sim}. Results are presented in Table \ref{tab:bysplit_results_splitsim_balanced}. 
When comparing {\sc 0-split} pairs to pairs involving split-words, we observe the same tendencies as in the full version of {\sc split-sim}: 
    for monosemous words and polysemous verbs, word splitting has a negative effect on word representations.
    There are, however, some differences in the significance of results, particularly in {\sc p-v}, due in part to the much smaller sample size of this dataset. 

It is important to note that split-types are strongly defined and determined by word frequency. In natural conditions (i.e., without controlling for frequency), we expect to encounter the patterns found in Table \ref{tab:bysplit_results_splitsim}.

\paragraph{The effect of $\Delta f$} 
   In Table \ref{tab:bysplit_results_splitsim_balanced}, we can 
   see that, in a dataset with lower and better balanced $\Delta f$ values, {\sc 1-split} pairs are no longer at a disadvantage and obtain results that are most of the time superior to those of {\sc 2-split} pairs. 
   We 
   run an additional analysis to study the effect of different $\Delta f$. 
   We divide the pairs in each subset and split-type according to whether their $\Delta f$ is below 
   or above 
   a 
   threshold $t=0.25$, 
   ensuring that all sets compared have at least 100 pairs. 
   Results, omitted for brevity, 
   show that pairs with lower $\Delta f$ obtain almost systematically 
   better results than those with higher $\Delta f$. 
   This confirms that a disparity in the frequency levels of the words compared also has a negative effect on similarity estimation.
   
\begin{table}[!t]
    \centering
    \scalebox{0.80}{
    \begin{tabular}{p{0.5cm}l|cc | cc | cc | cc}    
    & \multicolumn{1}{c|}{} & \multicolumn{2}{c|}{BERT} & \multicolumn{2}{c|}{CBERT} & \multicolumn{2}{c|}{ELECTRA} & \multicolumn{2}{c}{XLNet}  \\    
    & & L & H & L & H & L & H & L & H \\    
    \hline
    \multirow{3}{*}{{\sc m-n}} & {\sc 0-s} & 52 & 52 & 51 & 51 & \textbf{52} & 51 & \textbf{59} & 53 \\ 
          & {\sc 1-s} & 44 & \textbf{49} & 45 & \textbf{51} & 47 & \textbf{51} & \textbf{54} & 47 \\ 
    & {\sc 2-s} & \textbf{47} & 42 & \textbf{54} & 45 & \textbf{49} & 46 & \textbf{52} & 47 \\ 
    \hline
    \multirow{3}{*}{{\sc p-n}} & {\sc 0-s} & 36 & \textbf{43} & 38 & \textbf{40} & 39 & \textbf{40} & 40 & \textbf{42} \\ 
    & {\sc 1-s} & 45 & 45 & \textbf{47} & 44 & 45 & \textbf{48} & \textbf{51} & 37 \\ 
    & {\sc 2-s} & \textbf{43} & 42 & \textbf{52} & 40 & \textbf{50} & 45 & \textbf{48} & 43\\
    \hline
\multirow{3}{*}{{\sc p-v}} & {\sc 0-s} & \textbf{40} & 39 & \textbf{39} & 38  & \textbf{41} & 39 & \textbf{48} & 38 \\ 
    & {\sc 1-s} & \textbf{47} & 39 & 39 & \textbf{42} & \textbf{48} & 44 & \textbf{44} & 38 \\ 
    & {\sc 2-s} & 36 & \textbf{40} & \textbf{41} & 38 & 36 & \textbf{41} & \textbf{41} & 39\\     
    \hline
    \multicolumn{10}{c}{Without context}  \\
    \hline
    \multirow{3}{*}{{\sc m-n}} & {\sc 0-s} & 37 & \textbf{43} & 44  & \textbf{46} &  45 & \textbf{49} & \textbf{58} & 51 \\
          & {\sc 1-s} & 24 & \textbf{29} & 39 & \textbf{42} & 30 & \textbf{32} & 31 & \textbf{32} \\
    & {\sc 2-s} & \textbf{29} & 28 & 36 & 36 & \textbf{32} & 27 & \textbf{32} & 29 \\
    \hline
    \multirow{3}{*}{{\sc p-n}} & {\sc 0-s} & 14 & \textbf{29} & 32 & \textbf{36} & 19 & \textbf{36} & 33 & \textbf{34}  \\    
    & {\sc 1-s} & 25 & \textbf{27} & \textbf{40} & 35 & 25 & \textbf{32} & \textbf{28} & 23 \\
    & {\sc 2-s} & 22 & \textbf{28} & \textbf{35} & 34 & \textbf{26} & 25 & \textbf{23} & 21\\
    \hline
\multirow{3}{*}{{\sc p-v}} & {\sc 0-s} & 29 & \textbf{34} & 29 & \textbf{32} & 36 & \textbf{41} & \textbf{44} & 38\\
    & {\sc 1-s} & \textbf{34} & 18 & \textbf{41} & 33 & \textbf{29} & 21 & 23 & \textbf{25}\\
    & {\sc 2-s} & 19 & \textbf{27} & 31 & \textbf{34} & 15 & \textbf{31} & 22 & \textbf{25}\\      
    
    \end{tabular}}
    \caption{Results on pairs with low (L) and high (H) frequency using 10 (top) and no (bottom) contexts.}
    \label{tab:bysplit_results_LHfreq}
\end{table}

\paragraph{The effect of frequency on similarity estimation}

To investigate how estimation quality varies with frequency, we divide the data in every subset and split-type into two sets, L (low) and H (high), based on individually determined frequency thresholds. 
Using different thresholds does not allow us to 
fairly compare across data subsets and split-types but ensures that both classes (L and H) are always well-represented and balanced. 
The frequency of a word pair is calculated as the 
average frequency of the two words in it. To prevent L and H from containing pairs of similar frequency, their thresholds are apart by 0.25.
We only include pairs with a $\Delta f$ of at most 1. {\sc m-v} is excluded from this analysis because of its small size. 

Table \ref{tab:bysplit_results_LHfreq} (top section) shows results of this analysis.
Very often,
correlations are higher on the sets of pairs with lower average frequency (L). 
This is surprising, because, as explained in Section \ref{sec:background}, rare words are typically problematic in NLP. Works investigating the representation of rare words in BERT, however, either test it through prompting \cite{schick2020rare}, on ``rarified'' downstream tasks \cite{schick-schutze-2020-bertram}, or on word similarity but without providing contexts \cite{li-etal-2021-learning}. 
We believe the observed result is due to a combination of multiple factors; both contextual and lexical. First, the contexts used to extract representations provide information about the word's meaning. If we compare results to a setting where words are presented without context (lower part of Table \ref{tab:bysplit_results_LHfreq}), 
the tendency is indeed softened, 
but not completely reversed, meaning that context alone does not fully explain this result. 
Lower frequency words are also more often morphologically complex than higher frequency ones. This is the case in our dataset.\footnote{We verify this with the MorphoLEX \cite{sanchez2018morpholex} and LADEC \cite{gagne2019ladec} databases.} 
In the case of split-words,
morphological complexity may be an advantage that helps the model understand word meaning through word splitting. 
Another factor contributing to this result may be the degree of polysemy. 
We have seen in 
Table \ref{tab:bysplit_results_splitsim} 
that similarity estimation tends to be of better quality on monosemous words than on polysemous words. 
However, a definite 
explanation of the observed results would require additional analyses which are beyond the scope of this study.

\subsubsection{Further Analysis}

\begin{figure}[]
    \centering
    \includegraphics[width=\columnwidth]{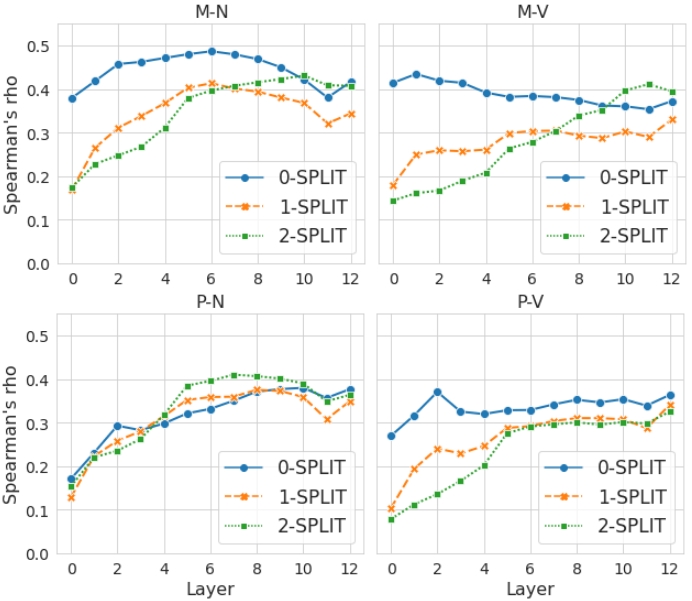}
    \caption{BERT {\tt AVG} results by layer and split-type on every {\sc split-sim} subset.} 
    \label{fig:by_layers_bert_avg}
\end{figure}

\paragraph{How do results change across layers for every split-type?}
Figure \ref{fig:by_layers_bert_avg} shows the BERT {\tt AVG} performance on each split-type of every subset 
across model layers. In {\sc m-n}, {\sc m-v} and {\sc p-v} we observe that at earlier layers the quality of the similarity estimations involving split-words is lower than that of {\sc 0-split} pairs. However, as information advances through the network and the context is processed, their quality 
improves at a higher rate than that of {\sc 0-split}, which remains more stable. This suggests that split-words 
benefit from the contextualization process taking place in the Transformer layers more than full-words. This makes sense, since 
sub-tokens 
are highly ambiguous (i.e., they can be part of multiple words), so more context processing is needed for the model to represent their meaning well. In a similar vein, the initial advantage of {\sc 0-split} pairs 
is more pronounced in monosemous words, which is expected as context is less crucial for understanding their meaning.  
In {\sc p-n}, the situation is different: {\sc 0-split} pairs behave in a similar way as 1- and {\sc 2-split} pairs from the very first layers. 
We verify whether this could be 
due to non-split polysemous nouns in {\sc p-n} being particularly ambiguous. We obtain their number of senses and we also check how many split-words in WordNet they are part of following BERT's tokenization (e.g., the word ``station'' is part of \{station, \#\#ery\}). These figures, however, are higher in {\sc p-v}, so this hypothesis is not confirmed. 

We also note that performance for the different split-types usually peaks at different layers. This highlights the need to carefully select the layer to use depending on the word's tokenization.

The same tendencies are observed with ELECTRA and XLNet. 
In CBERT, results are much more stable across layers.

\paragraph{Is a correct morphological segmentation important
for the representations' semantic content?} 
As explained in Section \ref{sec:background}, 
the morphological awareness of a tokenizer 
has a positive effect on results in NLP tasks. Here we verify 
whether 
it is also beneficial for word similarity prediction.
We use MorphoLex, 
a database containing morphological information (e.g., segmentation into roots and affixes) on 70,000 English words.
We consider that a split-word in {\sc split-sim} 
is incorrectly segmented if one or more of the roots of the word have been split
(e.g., \textit{saltshaker}: \{\textit{salts}, \textit{\#\#hak}, \textit{\#\#er}\}).\footnote{We do not base the definition of an incorrectly segmented word 
on the preservation of affixes because the segmentation in MorphoLex contains versions of affixes that do not always match the form realized in the word (e.g., sporadically = sporadic + ly).}
We compare the performance on word pairs involving an incorrectly segmented word ({\sc inc}) 
to that of pairs where the root(s) are fully preserved in both words ({\sc cor}), regardless of whether the tokens containing the root contain other affixes (e.g., \{\textit{marina}, \textit{\#\#te}\}). 
Note that MorphoLex does not fully cover the vocabulary in {\sc split-sim}.\footnote{Its coverage ranges between 38\% and 76\% of words depending on the subset.} We 
exclude {\sc m-v} from this analysis because of the insufficient amount of known {\sc cor} pairs (4 in {\sc 2-split} following BERT's tokenization). All other comparisons involve at least 149 pairs.
Results are presented in Table \ref{tab:morpho}. They confirm 
that, in subword-based models, when tokenization aligns with morphology, representations are almost always of better quality than when it does not.
The results obtained with CBERT, evaluated according to BERT's tokenization, highlight that the same set of {\sc inc} pairs is not necessarily harder to represent than {\sc cor} for a model that does not rely on subword tokenization.

\begin{table}[!t]
    \centering
    \scalebox{0.83}{
    \begin{tabular}{ll|cc|cc|cc}    
    & \multicolumn{1}{c|}{} & \multicolumn{2}{c|}{{\sc m-n}} & \multicolumn{2}{c|}{{\sc p-n}} & \multicolumn{2}{c}{{\sc p-v}} \\
    & & {\sc 1-s} & {\sc 2-s} & {\sc 1-s} & {\sc 2-s} & {\sc 1-s} & {\sc 2-s} \\ 
    \hline
    \multirow{2}{*}{BERT} & {\sc cor} & \textbf{47} & 35 & \textbf{39} & \textbf{52} & 34 & \textbf{48}  \\
     & {\sc inc} & 44  & \textbf{40} & 36 & 38 & \textbf{35}  & 32 \\
     \hline
     \multirow{2}{*}{CBERT} & {\sc cor} & 41 & 38 & 27 & 36 & \textbf{28} & \textbf{54} \\
     & {\sc inc} & \textbf{43} & \textbf{43} & \textbf{31} & \textbf{41} & 26 & 30 \\
     \hline
     \multirow{2}{*}{ELECTRA} & {\sc cor} & \textbf{46} & \textbf{51} & \textbf{41} & \textbf{57} & 28 & \textbf{50}\\     
     & {\sc inc} & 44 & 43 & 37 & 41 & \textbf{31} & 30 \\
     \hline
     
     \multirow{2}{*}{XLNET} & {\sc cor} & \textbf{52} & \textbf{58} & \textbf{40} & \textbf{51} & \textbf{42} & \textbf{44} \\
     & {\sc inc} & 47 & 43 & 38 & 42 & 35 & 33\\
    \end{tabular}}
    \caption{Spearman's $\rho$ ($\times$ 100) on pairs with an incorrectly segmented word ({\sc inc}) and pairs where the root(s) of both words are preserved ({\sc cor}).}
    \label{tab:morpho}
\end{table}

\paragraph{Do similarity predictions vary across split-types?} 

\begin{figure}[t!]
    \centering
    \includegraphics[width=\columnwidth]{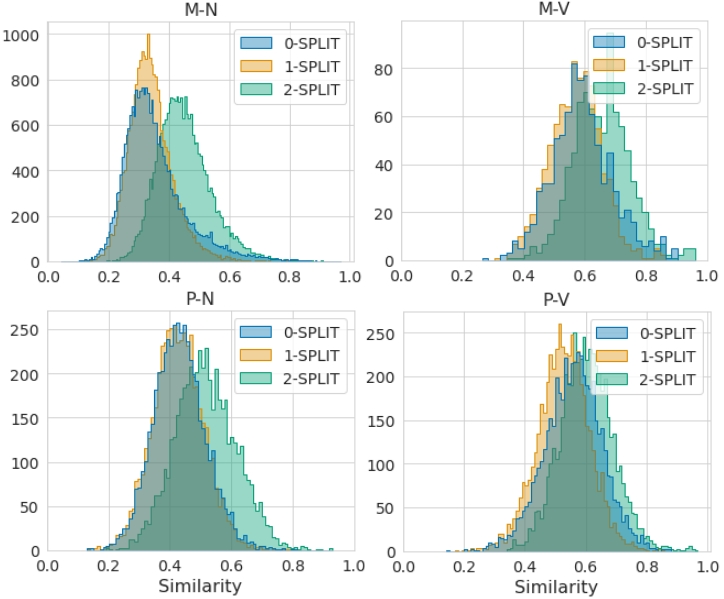}
    \caption{Distribution of predicted 
    similarity values by BERT ({\tt AVG}) across split-types in {\sc split-sim}.}
    \label{fig:prediction_histograms_wordnet}
\end{figure}

In Figure \ref{fig:prediction_histograms_wordnet} we show the histogram of similarities calculated with BERT {\tt AVG} using the best overall layer (cf. Table \ref{tab:global_results_splitsim}). We observe that similarity values 
are found in different, though overlapping, ranges depending on the split-type. {\sc 2-split} pairs exhibit a clearly higher average similarity than 0- and {\sc 1-split} pairs. Similarities in {\sc 1-split} tend to be the lowest, but the difference is smaller.
This does not correspond to 
the distribution of gold {\sc wup} similarities, which, due to our data collection process, does not differ across split-types. 
A possible partial explanation is that sub-token (\#\#) representations are generally closer together because they share distributional properties.\footnote{We indeed find that, in BERT's embedding layer, similarity between random sub-tokens is 
slightly higher (0.46) than between full-tokens or in mixed pairs (0.44 in both cases).} 
The same phenomenon is found in all models tested (ELECTRA, XLNet and CBERT), but is less pronounced in nouns in XLNET. 

This observation has important implications for similarity interpretation, and it discourages the comparison 
across split-types even when considering words of the same degree of polysemy and PoS. A similarity score that may be considered high for one split-type may be just average for another.

\begin{table}[!t]
    \centering
    \scalebox{0.87}{
    \begin{tabular}{ll|cc|cc|cc}    
    & \multicolumn{1}{c|}{} & \multicolumn{2}{c|}{BERT} & \multicolumn{2}{c|}{ELECTRA} & \multicolumn{2}{c}{XLNet} \\
    & & $-$ & $+$ & $-$ & $+$ & $-$ & $+$ \\ 
    \hline
    
\multirow{2}{*}{{\sc m-n}} & {\sc 1-s} & 41 & \textbf{42} & 42 & 42  & \textbf{48} & 45 \\ 
& {\sc 2-s} & 37 & \textbf{49} & 44 & \textbf{48} & 46 & \textbf{49} \\     
    \hline
    \multirow{2}{*}{{\sc m-v}} & {\sc 1-s} & 33 &  \textbf{36} & \textbf{37} & 36 & \textbf{37} & 32 \\    
    & {\sc 2-s} & 35 & \textbf{51} & 31 & \textbf{44} & 35 & \textbf{42} \\
    \hline   
    \multirow{2}{*}{{\sc p-n}} & {\sc 1-s} & \textbf{38} & 37 & \textbf{40} & 34 & 38 & \textbf{39} \\ 
    & {\sc 2-s} & 42 & 42 & \textbf{46} & 44 & \textbf{49} & 44 \\ 
    \hline
\multirow{2}{*}{{\sc p-v}} & {\sc 1-s} & 32 & \textbf{36} & 28 & \textbf{31} & 35 & 35  \\
    & {\sc 2-s} & \textbf{34} & 31 & \textbf{32} & 30 & 35 & \textbf{36} \\    
    \end{tabular}}
    \caption{Spearman's $\rho$ ($\times$ 100) obtained on {\sc split-sim} pairs tokenized into few ($-$) or many ($+$) subwords.}
    \label{tab:numpiece}
\end{table}

\paragraph{Does the number of subwords have an impact on the representations' semantic content?} 
We saw in Section \ref{sec:background} that oversplitting words has negative consequences on certain NLP tasks. We investigate the effect
that the number of subwords has on similarity predictions. 
We depart from the hypothesis that the more subwords a word is split into, the worse the performance will be. This is based on the intuition that shorter subwords are not able to encode as much lexical semantic information as longer ones. 
We count the total number of subwords
in each word pair and re-calculate correlations separately on sets of word pairs with few ($-$) or many ($+$) subwords.  
In {\sc 1-split}, ``$-$'' is defined as 3 subwords and in {\sc 2-split}, as 5 or less. We make sure that every set contains at least 1,000 pairs. 
Results 
are presented in Table \ref{tab:numpiece}. 
Our expectations are only met in about half of the cases, particularly in {\sc p-n}. Surprisingly, similarity estimations from BERT tend to be 
more accurate when words 
are split into a larger number of tokens, even though the tokenization in $+$ is more often morphologically incorrect than in $-$. 
Results from other models are mixed. 

Since only the first subword in a split-word is a full-token (i.e., does not begin with \#\# in BERT), one difference between words split into few or many pieces is the ratio of full-tokens to sub-tokens. 
When using the {\tt AVG} strategy, on ``$-$'' 
split-words, the first subword (a sub-token) has a large impact on the final representation, 
which is reduced as the number of subwords increases. We investigate whether this difference has something 
to do with the results obtained with BERT. 
To do so, we test two more word representation strategies: {\tt o1}, where we omit the first subword (the full-token) and {\tt oL}, where we omit the last subword (a sub-token). 
If 
mixing the two kinds of subwords (sub-tokens and full-tokens) is detrimental for the final representation, we expect {\tt o1} to obtain better results than {\tt oL}. 
Results by these two strategies could be affected by the 
morphological structure of words in {\sc split-sim} (e.g., {\tt o1} could perform better than {\tt oL} on words with a prefix). To control for this, 
we only run this analysis on word pairs consisting of two simplexes (according to MorphoLex). We exclude {\sc m-v} because of the insufficient ($< 100$) amount of pairs available in each class.

\begin{table}[!t]
    \centering
    \scalebox{0.9}{
    \begin{tabular}{ll|cc|cc|cc}    
    & \multicolumn{1}{c|}{} & \multicolumn{2}{c|}{{\sc m-n}} & \multicolumn{2}{c|}{{\sc p-n}} & \multicolumn{2}{c}{{\sc p-v}} \\
    & & {\sc 1-s} & {\sc 2-s} & {\sc 1-s} & {\sc 2-s} & {\sc 1-s} & {\sc 2-s}  \\   \hline    
\multirow{2}{*}{$-$} & {\tt o1} & \textbf{51} & \textbf{42} & \textbf{33} & 32 & \textbf{30} & 36 \\
& {\tt oL} & 42 & 37 & 30 & \textbf{33} & 28 & \textbf{38} \\   
\hline    
    \multirow{2}{*}{{$+$}} & {\tt o1} & \textbf{47} & \textbf{37} & \textbf{42} & 26 & 39 & \textbf{41}\\    
    & {\tt oL} & 43 & 29 & 40 & \textbf{45} & 39 & 37\\
    
    \end{tabular}}
    \caption{Results obtained with BERT {\tt AVG} omitting the first ({\tt o1}) or last ({\tt oL}) token on simplex {\sc split-sim} pairs tokenized into different amounts of subwords.}
    \label{tab:omitresults}
\end{table}

Results of this analysis are shown in Table \ref{tab:omitresults}.
In most cases, particularly in {\sc m-n}, the {\tt o1} strategy, which excludes the only full-token in the word, obtains a better performance than {\tt oL}.
This suggests that, in the BERT model, the first token is less useful 
when building a representation.
This is surprising, 
because English tends to place disambiguatory cues 
at the beginning of words \cite{pimentel-etal-2021-disambiguatory},  
 and because the first subword is often the longest one.\footnote{This is the case in 56 to 60\% of split-words in {\sc split-sim}, depending on the subset.} 
 The intuition that representations of longer tokens contain more semantic information is, thus, not confirmed.

\subsection{Within-Word} \label{sec:withinword_results}

\begin{figure}
    \centering
    \includegraphics[width=\columnwidth]{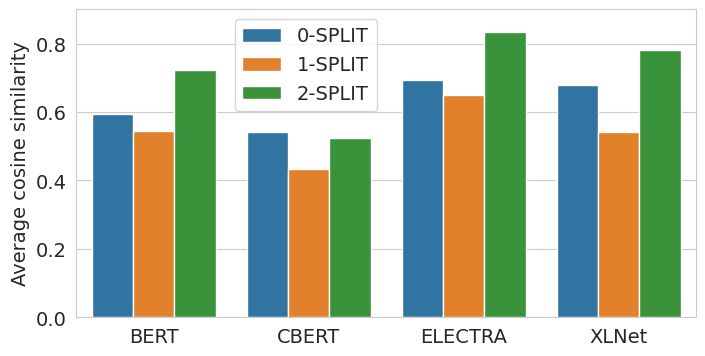}
    \caption{Average similarity values obtained on WiC ({\sc bal}) with the {\tt AVG} strategy.}
    \label{fig:wic_similarities}
\end{figure}

In this section we present the results 
on the WiC dataset. In 
Table \ref{tab:wic_results}, we report the best accuracy obtained by every model on different split-types. 
We observe that the best performance is achieved on the full set of {\sc 2-split} pairs ({\sc all}). 
This can be explained by the label distribution in {\sc 2-split}, where most pairs are of type T (cf. Table \ref{tab:wic_stats}). We have seen in Section \ref{sec:interword_results} that 
{\tt AVG} representations for these pairs have higher similarity values, and we 
confirm 
this is the case, too, in the within-word setting (see Figure \ref{fig:wic_similarities}). In fact, in the case of BERT {\tt AVG}, 
only 18 out of 97 F 
{\sc 2-split} word pairs were correctly guessed. To have a fairer comparison with 
{\sc 0-split} pairs, where labels are more balanced, 
we recalculate accuracy on {\sc 1-} and {\sc 2-split} pairs 
randomly subsampling as many T pairs as the number of available F pairs ({\sc bal}). 
These results are shown in the same Table.
From them, we conclude that accuracy on {\sc 1-} and {\sc 2-split} pairs is actually lower than that on {\sc 0-split}. 
This is not true of CBERT, however, which performs equally well across split-types and is 
the best option for {\sc 2-split} pairs.  As we can see in Figure \ref{fig:wic_similarities}, 
the similarities it assigns to {\sc 2-split} are in
a similar range to {\sc 0-split} in this within-word setting.

\begin{table}[!t]
    \centering
    \scalebox{0.82}{
    \begin{tabular}{ll|c|cc|cc}    
    & \multicolumn{1}{c|}{} & \multicolumn{1}{c|}{{\sc 0-s}} & \multicolumn{2}{c|}{{\sc 1-s}} & \multicolumn{2}{c}{{\sc 2-s}} \\
    & & {\sc all} &  {\sc all} & {\sc bal} & {\sc all} & {\sc bal} \\ 
    \hline
    \multirow{3}{*}{BERT} & {\tt AVG} & \multirow{3}{*}{70} & 66 & \textbf{67} & 75 & 57 \\
     & {\tt WAVG} &  & 65 & 63 & 75 & 58  \\
     & {\tt LNG} &  & 65 & 62 & 74 & \textbf{60} \\     
     \hline
     \multirow{3}{*}{FLOTA} & {\tt AVG} & \multirow{3}{*}{69} & 60 & \textbf{62} & 74 & \textbf{60}  \\
     & {\tt WAVG} & & 60 & 58 & 75 & 59 \\
     & {\tt LNG} & & 60 & 57 & 73 & \textbf{60} \\     
     \hline
     \multirow{1}{*}{CBERT} & - & 67 & 57 & 67 & 66 & 66\\          
     \hline
     \multirow{3}{*}{ELECTRA} & {\tt AVG} & \multirow{3}{*}{71} & 62 & \textbf{62} & 76 & 58 \\
     & {\tt WAVG} & & 62 & 59 & 76 & 61 \\
     & {\tt LNG} & & 57 & 59 & 75 & \textbf{65}  \\     
     \hline     
     \multirow{3}{*}{XLNET} & {\tt AVG} & \multirow{3}{*}{62} & 61 & 61 & 68 & \textbf{58} \\
     & {\tt WAVG} & & 62 & \textbf{62} & 69 & \textbf{58} \\
     & {\tt LNG} & & 62 & \textbf{62} & 68 & 57 \\     
    \end{tabular}}
    \caption{Accuracy obtained 
    on WiC on the full subsets ({\sc all}) and balancing T/F labels in {\sc 1-} and {\sc 2-split} ({\sc bal}). The best result per model and split-type in {\sc bal} subsets is boldfaced.}
    \label{tab:wic_results}
\end{table}

When it comes to the pooling strategy for representing split-words, 
{\tt AVG} is still often the best one, but {\tt LNG} also obtains good results. 
When comparing instances of the same word, contextual information is more important than word identity, so omitting part of a word does not have such a negative impact as in the inter-word setting.

In Table \ref{tab:wic_results_samediff}, 
we look at the results of {\tt AVG} on the original data and when replacing target words with their lemmas ({\tt LM}) separately on {\sc same} vs {\sc diff} pairs.
There is a large gap in accuracy between {\sc same} and {\sc diff} {\sc 2-split} pairs, with {\sc diff} pairs obtaining worse results with all models tested\footnote{A partial explanation is that {\sc same} pairs have a slightly stronger tendency of being T (77\% of {\sc same} {\sc 2-split} pairs are T, against 66\% of {\sc diff 2-split} pairs).} except XLNet. 
{\sc 0-split} pairs, on the contrary, are generally less 
affected by this parameter. 
While using the lemma 
is clearly helpful for  {\sc 1-split} pairs, it does not show a consistent pattern of improvement in the other split-types. 
We also observe that the average similarities for {\sc same} pairs are higher than for {\sc diff} pairs (e.g., BERT in {\sc 0-split}: 0.62 ({\sc same}), 0.54 ({\sc diff})).

\begin{table}[!t]
    \centering
    \scalebox{0.7}{
    \begin{tabular}{ll|cc|cc|cc|cc}    
    & \multicolumn{1}{c|}{} & \multicolumn{2}{c|}{BERT} & \multicolumn{2}{c|}{CBERT} & \multicolumn{2}{c|}{ELECTRA} & \multicolumn{2}{c}{XLNet} \\
    & & {\tt AVG} & {\tt LM} &{\tt AVG} & {\tt LM}& {\tt AVG} & {\tt LM} &  {\tt AVG} & {\tt LM} \\
    \hline    
\multirow{2}{*}{{\sc 0-s}} & {\sc same} & 70 & \textbf{73} & 67 & \textbf{68} & 71 & 71 &  64 & 64\\ 
& {\sc diff}  & \textbf{69} & 65 & \textbf{68} &  67 & \textbf{77} & 70 & 60 & \textbf{62}\\    
 \hline
 \multirow{2}{*}{{\sc 1-s}} & {\sc same} & - & - & - & - & - & - & 58 & \textbf{59} \\
 & {\sc diff} & 66 & \textbf{70} & 57 & \textbf{65} & 62 & \textbf{64} & 63 & \textbf{65} \\   
 \hline    
  \multirow{2}{*}{{\sc 2-s}} & {\sc same} & 79 & \textbf{80} & \textbf{73} & 69 & 80 & 80 & 68 & 67\\  
 & {\sc diff}  & \textbf{65} & 62 & 58 & \textbf{60} & \textbf{64} & 63 & 73 & 73\\   
    \end{tabular}}
    \caption{Accuracy on WiC pairs with the {\sc same} vs {\sc diff} surface form.}
    \label{tab:wic_results_samediff}
\end{table}

\section{Discussion}

We have seen that when examined separately, 
word pairs involving split-words often obtain worse quality similarity estimations than those consisting of full-words; but 
this depends on the type of word: split polysemous nouns are better represented 
than non-split ones. This holds across the models and tokenizers tested, and also when evaluating on words in a narrower frequency range. This shows that word splitting has a negative effect on the representation of many words.
We have also seen 
that in normal conditions, 
performance on {\sc 1-split} is generally 
the worst one, due mainly to a larger disparity in frequencies of the words in a pair.
Our analysis has also confirmed the hypothesis that words that are split in a way that preserves their morphology obtain better quality similarity estimates than words where segmentation splits the word's root(s).

We have noted that similarities for the different split-types are found in different ranges; notably, similarities between two split-words
tend to be higher than similarities in {\sc 0-} and {\sc 1-split} pairs. Naturally, this has an effect on the correlation calculated on the full dataset, which is lower than when considering each split-type separately.
It would be interesting to develop a similarity measure that allows comparison across split-types, which could rely on information from the rest of the sentence, like 
BERTScore \cite{zhang2020bertscore}. 
Another simple way to make similarities comparable would be to bring {\sc 2-split} similarities to the {\sc 0-split} similarity range by subtracting the average similarity value obtained in {\sc 0-split}. The best value to use, however, may vary depending on the application.

One surprising finding relates to the impact of the number of subwords: similarity estimations are not always more reliable on words involving fewer tokens. This was especially the case for BERT, where we saw that the first token is generally the least useful in building a representation. Given the tendency for the first token to be the longest, this has put the other strategies tested ({\tt WAVG} and {\tt LNG}) at a disadvantage.

From our within-word experiments we confirm that word form is reflected in the representations and has a strong impact on similarity, but this does not necessarily mean 
that comparing words with distinct morphological properties 
(e.g., singular vs plural) would be detrimental in the inter-word setting.
In the within-word setting, 
{\sc same} pairs compare two equal word forms, whose representation at the initial (static) 
embedding layer is identical. {\sc diff} pairs, instead, start off with different static embeddings, which results in an overall lower similarity. 
In {\sc split-sim}, all comparisons are made, by definition, between different words. 
The fact that two words have different morphological properties may thus have a smaller impact on results. 

Most of our findings are consistent between the two kinds of task (inter- and within-word) and across models. One exception is CBERT, which does not assign higher similarities to {\sc 2-split} pairs when comparing instances of the same word; and the {\tt LNG} strategy, which is more useful within-word than inter-word. {\tt AVG} is however the best strategy overall. One direction for future work would be to find 
a pooling method that closes the gap in performance between split-types.

Our experiments only involve one language (English), Spearman's correlation and cosine similarity, although our methodology is not restricted to a single similarity or evaluation metric. Extending this work to more languages is also possible, but less straightforward, due to the need for suitable datasets.

\section{Conclusion}

We have compared the contextualized representations of words that are segmented into subwords to those of words that have 
a dedicated embedding in BERT and other models. We have done so 
through an intrinsic evaluation relying on similarity estimation.
Our findings are relevant for any NLP practitioner 
working with contextualized word representations, and particularly for applications relying on word similarity:  
(i) Out of the tested strategies for split-word representation, averaging subword embeddings is 
the best one, with few exceptions;
(ii) the quality of split-word representations is  often worse than that of full-words, although this depends on the kind of words considered;
(iii) similarity values obtained for split-word pairs are generally higher than similarity estimations involving full-words; 
(iv) the best layers to use differ across split-types;
(v) 
a higher number of tokens 
does not necessarily, as intuitively thought, decrease representation quality;
(vi) in the within-word setting, word form has a 
negative impact on results when words are split. 

Our results also point to specific aspects 
to which future research and efforts of improvement should be directed. We make our {\sc split-sim} dataset available to facilitate 
research on split-word representation.

\section*{Acknowledgments}

We thank the anonymous reviewers and the TACL Action Editor for their thorough reviews and useful remarks, which helped improve this paper. This research has been supported by the Télécom Paris research chair on Data Science and Artificial Intelligence for Digitalized Industry and Services (DSAIDIS) and by the Agence Nationale de la Recherche, REVITALISE project (ANR-21-CE33-0016).

\bibliography{anthology,custom}
\bibliographystyle{acl_natbib}

\appendix

\begin{table*}[!t]
    \centering
    \scalebox{0.87}{
    \begin{tabular}{ll|ccc | ccc | c | ccc | ccc}    
    & \multicolumn{1}{c|}{} & \multicolumn{3}{c|}{BERT} & \multicolumn{3}{c|}{BERT-FLOTA} & \multicolumn{1}{c|}{CBERT} & \multicolumn{3}{c|}{ELECTRA} & \multicolumn{3}{c}{XLNet}  \\
    & & {\tt AVG} & {\tt WAVG} & {\tt LNG} & {\tt AVG} & {\tt WAVG} & {\tt LNG} & - & {\tt AVG} & {\tt WAVG} & {\tt LNG} & {\tt AVG} & {\tt WAVG} & {\tt LNG} \\
    \hline
    \multirow{3}{*}{{\sc m-n}} & {\sc 0-s} & \multicolumn{3}{c|}{65} & \multicolumn{3}{c|}{65} & 68 & \multicolumn{3}{c|}{68} & \multicolumn{3}{c}{\underline{74}} \\ 
    \cdashline{3-15}
    & {\sc 1-s} & \textbf{45}* & 43* & 32* & \textbf{39}* & 37* & 31* & 55* & \textbf{50}* &  49* & 42* & \textbf{\underline{59}}* & 58* & 56* \\ 
    & {\sc 2-s} & \textbf{46}* & 40* & 29* & \textbf{36}* & \textbf{27}* & 25* & \underline{50}* & \textbf{47}* & 46* & 32* & \textbf{\underline{50}}* & 49* & 43* \\ 
    \hline
\multirow{3}{*}{{\sc m-v}} & {\sc 0-s} & \multicolumn{3}{c|}{66} & \multicolumn{3}{c|}{66} & 65 & \multicolumn{3}{c|}{70} & \multicolumn{3}{c}{\underline{74}} \\    
    \cdashline{3-15}
    & {\sc 1-s} & \textbf{45}* & 44* & 36* & \textbf{33}* & 33* & 29* & 52* & \textbf{51}* &  50* & 43* & \underline{\textbf{55}}* & 54* & 54* \\
    & {\sc 2-s} & \textbf{53}* & 50* & 38* & \textbf{37}* & \textbf{34}* & 26* & \underline{55}* & \textbf{53}* & 51* & 39* & \textbf{51}* & 50* & 48* \\ 
    \hline
    \multirow{3}{*}{{\sc p-n}} & {\sc 0-s} & \multicolumn{3}{c|}{52} & \multicolumn{3}{c|}{52} & 56 & \multicolumn{3}{c|}{54} & \multicolumn{3}{c}{\underline{60}} \\ 
\cdashline{3-15}
    & {\sc 1-s} & \textbf{40}* & 39* & 30* & \textbf{33}* & 31* & 25* & 50* & \textbf{47}* & \textbf{48}* & 41* & \textbf{\underline{52}} & \textbf{\underline{52}} & 50 \\ 
    & {\sc 2-s} & \textbf{49}* & 48* & 32* & 36* & \textbf{36}* & 27* & \underline{57} & 52 & \textbf{53} & 38* & \textbf{52} & \textbf{52} & 46 \\ 
\hline
\multirow{3}{*}{{\sc p-v}} & {\sc 0-s} & \multicolumn{3}{c|}{63} & \multicolumn{3}{c|}{63} & 56 & \multicolumn{3}{c|}{64} & \multicolumn{3}{c}{\underline{66}} \\
\cdashline{3-15}
    & {\sc 1-s} & \textbf{46}* & 45* & 37* & \textbf{37}* & 36* & 29* & 54 & \textbf{47}* & \textbf{47}* & 40* & \textbf{\underline{55}}* & \textbf{\underline{55}}* & 53* \\ 
    & {\sc 2-s} & \textbf{\underline{52}}* & 51* & 38* & 30* & \textbf{31}* & 25* & \underline{52}* & 51* & \underline{\textbf{52}}* & 38* & 50* & \underline{\textbf{52}}* & 47* \\      
    \end{tabular}}
    \caption{Spearman's $\rho$ ($\times$ 100) 
    on 
    {\sc split-sim} (full) using cosine similarities from FastText as a reference.} 
\label{tab:bysplit_results_splitsim_fasttext}
\end{table*}

\section{Results with FastText} \label{app:fasttext}

We choose FastText as a control because of its good results on word similarity, 
and because it can generate embeddings for all words. 
91.8\% of all pairs in {\sc split-sim} have both words present in the FastText vocabulary.\footnote{The class that is least well represented is {\sc 2-split} {\sc m-n}, but it still has a large majority of in-vocabulary words, with 79\% of pairs being completely covered.}
Table \ref{tab:bysplit_results_splitsim_fasttext} contains the results. 
The main tendencies observed in Sections \ref{sec:best_strategy} and \ref{sec:isperformanceworse} are found in these results too: {\tt AVG} is the best overall strategy 
and predictions on {\sc 1-} and {\sc 2-split} pairs are almost consistently of lower quality than on {\sc 0-split} pairs. 
We also observe a couple of discrepancies with respect to {\sc wup}: correlations are higher overall, which makes sense as FastText is also a model that learns representations from text and all models (including FastText) have been trained on Wikipedia data. 
Another important difference is the relative performance of {\sc 0-split} and {\sc 2-split} in {\sc p-n}. While with {\sc wup} {\sc p-n} is the only dataset where splitting words is not detrimental to similarity estimation, this is not the case with FastText. 
However, we note that the difference in performance between {\sc 0-split} and {\sc 2-split} is much smaller in PN than in the other subsets. This shows that, also in this setting, split polysemous nouns have an advantage with respect to split-words of other types.

\end{document}